\documentclass[runningheads,a4paper]{llncs}

\usepackage{amssymb}
\usepackage{amsmath}
\usepackage{graphicx}
\usepackage[utf8]{inputenc}
\usepackage[T1]{fontenc}
\usepackage{hyperref}
\usepackage{array}
\usepackage{url}
\urldef{\mailsa}\path|{sbeliga, mpobar, smarti}@uniri.hr|    
\newcommand{\keywords}[1]{\par\addvspace\baselineskip
\noindent\keywordname\enspace\ignorespaces#1}

\begin{document}

\mainmatter  

\title{Normalization of Non-Standard Words in Croatian Texts}

\titlerunning{Normalization of Non-Standard Words in Croatian Texts}

%
%
\author{Slobodan Beliga, Miran Pobar and Sanda Martinčić-Ipšić}
\authorrunning{Normalization of Non-Standard Words in Croatian Texts}

\institute{Department of Informatics\\
University of Rijeka\\
Omladinska 14, 51000, Rijeka, Croatia\\
\mailsa\\
}

%
%

\toctitle{Normalization of Non-Standard Words in Croatian Texts}
\tocauthor{Slobodan Beliga, Miran Pobar, Sanda Martinčić-Ipšić}
\maketitle

\begin{abstract}
This paper presents text normalization which is an integral part of any text-to-speech synthesis system. Text normalization is a set of methods with a task to write non-standard words, like numbers, dates, times, abbreviations, acronyms and the most common symbols, in their full expanded form are presented. The whole taxonomy for classification of non-standard words in Croatian language together with rule-based normalization methods combined with a lookup dictionary are proposed. Achieved token rate for normalization of Croatian texts is 95\%, where 80\% of expanded words are in correct morphological form. 
\end{abstract}


\section{Introduction} 
Systems for speech synthesis carry out the conversion of arbitrary input text into synthesized speech \cite{7}. These systems consist of different components which enable speech generation. One of the components of a TTS system is text normalization that transforms non-standard text elements into their expanded form, preparing them for further processing in the system (g2p conversion, prosody generation, etc.). In most cases text normalization includes numbers, dates, time, abbreviations, acronyms, different symbols, currency, measurement units etc. 

First problem in text normalization is detection of non-standard words (NSW). Sometimes standard words and NSWs share the same written form \emph{pol} -\emph{North Pole, half} and \emph{pol. - /political/}. The second problem is writing the detected NSWs in their full expanded form. For example, the abbreviation '\emph{st.}' has to be written as '\emph{stoljeće}' (century) or as '\emph{student}' depending on the context.  

Common methods \cite{12} for speech normalization are: hand-written rule-based methods, lookup dictionary based method which uses predefined dictionary for normalization or semiautomatic approach which automatically expands a novel abbreviation. Solutions for the text normalization for English \cite{9,10,12}, French \cite{5}, Russian \cite{13}, Polish \cite{6}, German \cite{2}, Slovenian \cite{8}, Czech \cite{16} have been reported. 

The main aim of this article is to present the problems of Croatian text normalization and to propose algorithms for the normalization. The algorithms are presented according to the proposed taxonomy of Croatian NSW and implemented in Perl. The normalization results are presented and some ideas for future work are stated. The paper concludes with discussion on possible integration of proposed text normalization into the existing grapheme-to-phoneme conversion \cite{14} and speech generation modules of Croatian TTS synthesis systems \cite{15}.


\section{Text Normalization}

Normalization is the first step in the text pre-processing of TTS [7]. The normalization module is responsible for the identification of a single NSW token and for its transformation into expanded form. Usually NSWs are not listed in the dictionary and there is no unique rule for their expansion or pronunciation [12]. Further, they are more ambiguous than standard words in meaning or pronunciation. The first problem is to identify all NSWs in Croatian and separate it from standard words. The second problem is the transformation of detected NSW into expanded form, suitable for the TTS system. These two problems are the most obvious, but there are still some other issues to consider. For instance, when is a punctuation mark an end of a sentence, and when is an abbreviation? For example  in the sentence '\emph{Ivo je na natjecanju bio 3. i odlikovan je broncom.}' we would read the number /$3.$/ as 'the third'. But, how can the computer recognize this sentence as one and not as two sentences?



\section{Taxonomy of Normalization for Croatian Language}
In most TTS systems text normalization is accomplished by using hand-written rules that are defined for particular domains of application \cite{1,12}. Along with rules \cite{16}, n-gram models \cite{7,12}, decision trees and weighted finite-state transducers \cite{12} or lookup dictionaries (lists) of most frequent NSWs with their expanded form \cite{12} have been used. Listing all NSWs is tiring and it never ensures complete success of normalization, i.e. it does not guarantee that some novel NSW from the input text is also listed in the lookup dictionary. Therefore, we suggest the taxonomy that classifies all posible NSWs and therefore provides the complete framework for text normalization problem for Croatian language.

The module initially classifies NSWs as letters, numerals or combination as shown in the main classification tree in Fig. 1. It is important that the input texts are written strictly according to orthographic rules, since the suggested taxonomy is based on Croatian orthographic and grammatical rules. As we can see the classification tree branches: into the left tree for characters (Fig. 1), into the middle tree for digits (Fig. 2) and into the right tree for combined alphanumeric characters (Fig. 2). NSW usually doesn't carry information by which we could easily interpret and expand it to the correct branch of the tree. So we try to classify NSWs of same characteristics in one unique class. 
\begin{figure}
\centering
\includegraphics[scale=0.5]{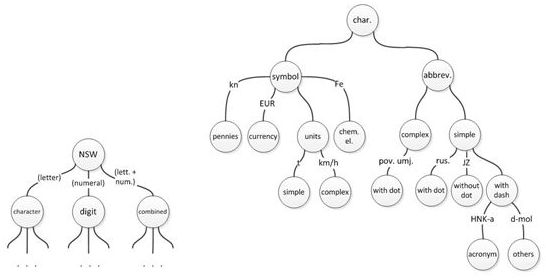}
\caption{Classification tree: main(left), characters(right).}
\end{figure}
With suggested classification it is possible to retrieve algorithms that make normalization more achievable and for certain classes the unified normalization algorithms are constructed. Instances within the certain class share common characteristics, like ordinal and cardinal numbers. Hence, algorithms for numbers normalization can be also applied to telephone numbers, dates and time. The Roman numbers are easily confused with letters (such as /\emph{Ivan Pavao II.}/ and /\emph{čl. II.}/) but they are expanded the same way as ordinal numbers.

\begin{figure}
\begin{centering}
\includegraphics[scale=0.45]{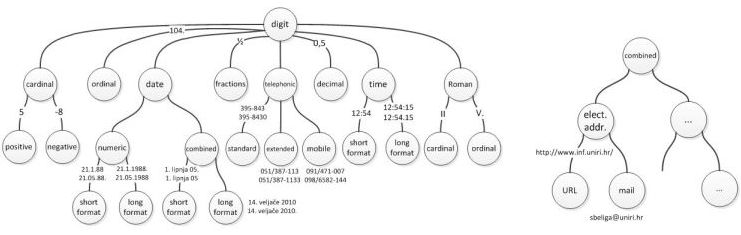}
\par\end{centering}

\caption{Classification tree: numbers(left), combined charecters(right).}

\end{figure}
Abbreviations, acronyms, symbols, measurement units and similar NSWs can apear in numerous forms and carry the meaning depending on the domain contex. Each field in science, culture or and society uses colloquial language, and its own NSWs. It is common that NSWs have more than one meaning and consequently more then one normalization form, based on the context. The normalization of abbreviations is carried out by combination of lookup lists (dictionaries) and some rules. Particularly complex group of NSW represent mixed semiotic sequences composed out of numbers and letters. Commonly they appear in IT related texts (e-mails \emph{\underline{ana.anic5@uniri.hr}} or url-s \href{http://perldoc.perl.org/per- fag5 htm}{\emph{http://perldoc.perl.org/per- fag5.htm}}).


\section{Implementation}
The normalization algorithms are implemented in Perl. Perl is suitable for text processing [11], because of many automated functions for solving problems of lexical analysis and functions for text processing. The proposed solution is based on the identification of NSW by using regular expressions which classify the NSW into correct class of the tree and writing it as an unambiguously pronaunceable text.

Numbers are highly suitable for normalization, because it is easy to determine how many ones, tens, hundreds, thousands they contain by the number of digits. By dividing them consecutively, we get the numeric value for each place.Then each digit is replaced with a word. Detailed review of number characteristics is given in \cite{4} as well as the detailed description of algorithms for normalization of ordinal and cardinal numbers. 

The numbers repeat the same pattern after every three digits. This fact implies that numbers can be normalized in a group of three digits with common characteristics. Once written, functions for expanding numbers of lower decadal place can be applied to upper places as well. Algorithms for each number NSW are based on the principle: search the root base according to the numeric values of decadal places, then add suffix, which is determined by the values of lower places. 

The ordinal number /\emph{21.}/ belongs to the interval $\left[11,100\right\rangle$. Each digit is decomposed according to its position and replaced by a written word: /\emph{21.}/ is replaced by /\emph{dva+deset i prvi}/. On the root base /\emph{dva}/ we add suffix /\emph{deset}/ and with the conjunction /\emph{i}/ add the word /\emph{prvi}/. Using the same principle we get the expanded forms of the ordinal numbers. Normalization of cardinal numbers is carried out by the same principle, except the used suffixes are different.

The rule of writing a dot after ordinal number is orthographically correct. But, sometimes a year without the dot at the end in written. Such incorrect form of writing is taken into consideration because many texts in newspapers, on web portals and in various documents generated by computer contain the years written without a dot at the end. Likewise, it is not necessary to write zero in front of single-digit numbers of days or months. 

The normalized form of date in Croatian standard language is written in nominative, except for the month which is always written in genitive. The time normalization is based on the same principle as cardinal numbers, only the intervals of $\left[0,60\right]$ are considered and  modified suffixes are used.

The normalization of abbreviations is carried out by lookup lists of the most frequently used abbreviations with their associated expanded forms. There are few common characteristics of abbreviations according to which we would formulate unique rule for normalization. For ambiguous NSWs it is necessary to use additional rule. For example, if the abbreviation /\emph{g.}/ comes after the ordinal number, it signifies a year /\emph{godina}/, after cardinal is /\emph{gram}/ and before or after a proper name can be /\emph{gospodin}/.

Acronyms are the subclasses of abbreviations. We suggest the following solution for acronyms: we detect them as  NSW tokens, but we never write them in their full expanded form, rather we write them as they are spelt. The acronym /\emph{MMF}/ is written as /\emph{ememef}/. This is also a suitable solution for foreign and ambiguous acronyms, as an example /\emph{DVD}/ is written as /\emph{devede}/ and the user of the system judges the true meaning depending on the context:  /\emph{Digital Video Disc}/ or  /\emph{Dobrovoljno vatrogasno društvo}/.


\section{Results and Integration}
The result of proposed Croatian text normalization are presented in terms of token correctness calculated as percentage of recognized tokens (the number of identified NSW in original text) divided with total tokens number (the number of total NSW in original text). Similar measure (token error rate) has been proposed in \cite{13}. Moreover, the text normalization in Croatian is complex due to the nature of Croatian language. Croatian is a highly flective Slavic language and words can have 7 different cases for singular and 7 for plural, genders and numbers. The measure of morphological correctness for evaluation of normalized NSWs in Croatian text was set as the flective correctness. It is the percentage of morphologically correct tokens out of correctly recognized tokens (the number of identified NSWs in original text).

The performance of Croatian text normalization was tested on the corpus of selected Croatian texts. The text collection included 18 texts with 11K words as shown in Table 1. Total number of NSWs in test text is 1728. The proposed Croatian text normalization correctly detected 1648 tokens, which resulted with 95,37\% overall token correctness. Among recognized tokens 1316 were in correct flective form which resulted with 80\% flective correctness. The test texts were collected according to their genre: educational, scientific, popular, news and formal. The text topics were: chemistry, physics, history, recipes, ads, weather reports, TV schedule, telephone directory for individuals and companies, road and travel conditions, law and legislation, political and election reports, business, exchange rates and currencies, etc. Fig. 3 presents the token and flective correctness calculated per each text genre. 
\begin{table}
\caption{The results per text genre.}
\begin{tabular}{|r|p{1.1cm}|p{1.1cm}|p{1.1cm}|p{1.3cm}|p{1.3cm}|p{1.8cm}|p{1.8cm}|}
\hline
Number of &	texts&	words&	total tokens&	unrecog. tokens&	recog. tokens &	correct tokens (morpho.)&	incorr. tokens (morpho.)\\
\hline\hline
Educational&	4  & 2714  &	272         &	24             &	248          &	92                      &	156\\
\hline
Scientific &	2	 & 543	 &  127         &  4	           &  123          &  81                      &	46\\
\hline
Popular	   &  3  & 633	 &  94	        &  1             &	93           & 	50                      &	43\\
\hline
News       &	5	 & 5814	 &  982         &  2	           &  980	         &  933	                    & 47\\
\hline
Formal     &	4  & 1230  & 	253         &	49	           &  204	         &  160	                    & 86\\
\hline \hline
\textbf{OVERALL}&18 &	10934 &  1728     &	80	           &  1648	       &  1316	                  &378\\
\hline 
\end{tabular}
\end{table}

The problems with abbreviations, symbols, measurement units, acronyms, etc. arise because they sometimes change depending on the context in which they are written and therefore it is very difficult to unify them in one algorithm. Additionally sometimes they share the same graphical format with standard words: /\emph{Na}/ can be chemical element /\emph{natrij}/ (sodium) or preposition /\emph{na}/ (on), /\emph{C}/ can be either \emph{Coulomb, carbon} or simply the home number 5C in address etc. For instance in presented test texts abbreviation /\emph{st.}/ was expanded as \emph{student, century, senior, item} and \emph{saint} depending on the text genre, which was determined in advance. 

Correct interpretation of semantic context in which NSW appears is an important field for our future research. Each scientific, cultural or social area uses its own colloquial language. For this reason, the dictionaries of frequently used abbreviations and symbols should be adapted to limited linguistic domain that the system will use. The language is constantly changing and evolving. Consequently, for the purposes of speech synthesis novelties in language should expand and update the normalization module in order to keep pace with stratified time, space and functionality. The second focus of our research will be morphological generation of correct flective form of normalized words, according to morphosyntactic tags of neighboring words.  
\begin{figure}
\begin{centering}
\includegraphics[scale=0.5]{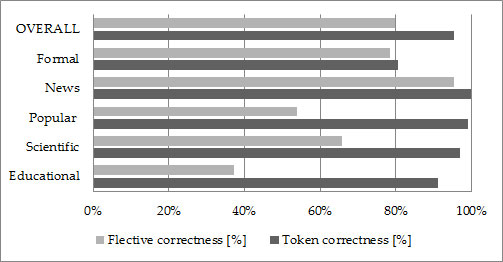}
\par\end{centering}

\caption{Token and flective correctness per text genre.}

\end{figure}

The proposed normalization can be easily integrated with existing grapheme-to-phoneme conversion \cite{14} and speech generation modules of TTS system \cite{15} which is under development. The fully integrated TTS system for Croatian can be used for applications in assistive technology, spoken information retrieval or simply as the reader.


\section{Conclusion}
This paper describes the normalization of non-standard words in Croatian texts for the purposes of speech synthesis. The text normalization is highly complex if we take into consideration the determination of correct gender, number and case of the normalized words. The problem is also the fact that input texts are not entirely written according to the orthographic principles of the Croatian standard language and the module for text normalization has to possess a certain degree of tolerance in conducting its methods that make the systems even more complex. 
The synthesized text is better and more complete if as many samples as possible are expanded in the process of text pre-processing. For that purpose, we suggested the taxonomy of Croatian NSWs which unifies the normalization procedures of Croatian texts. Algorithms for detection of samples for normalization (ordinal and cardinal numbers, dates of numeral and combined forms, time, abbreviations, acronyms and symbols) and algorithms for the normalization of identified forms were constructed as the combination of programmed rules and lookup dictionary in Perl. The proposed algorithms were tested on 18 texts of different genres: educational, scientific, popular, news and formal and overall token rate of 95\% and overall 80\% of correct flective forms were achieved. Integration of proposed text normalization into the existing grapheme-to-phoneme conversion \cite{14} and speech generation modules of TTS synthesis system [15] is under development. 

The language is constantly changing so some further efforts should be invested in continuous gathering of Croatian text, with different topics and discourse for keeping the normalization procedure up to date. Further, the interpretation of semantic context in which NSW appears should be addressed in future research. And finally, the normalization should generate the correct morphological form of expanded word.


\begin{thebibliography}{100}
\bibitem{1} Allen, J., Hunnicutt, M.S., Klatt, D.: From Text to Speech: the MITalk System. Cambridge University Press (1987)
\bibitem{2} Mobius, B. et al.: The Bell Labs German Text-to-Speech System: An Overview. EUROSPEECH 97, pp. 2443-2446. (1997)
\bibitem{3} Babić, S., Finka, B., Moguš, M.: Hrvatski pravopis, Školska knjiga, Zagreb (1996)
\bibitem{4} Beliga, S.: Normalizacija brojeva i datuma u postupcima umjetne tvorbe Hrvatskoga govora, Završni rad, Sveučilište u Rijeci, Odjel za informatiku, Rijeka (2010) 
\bibitem{5}	Larreur, D., Emerard, F., Marty, F.: Linguistic and Prosodic Processing for a Text-to-Speech Synthesis System, Eurospeech89, Paris, pp. 510-513 (1989)
\bibitem{6} Oliver, D.: Polish Text-to-Speech Synthesis. M.Sc. thesis, Univ. of Edinburgh (1998)
\bibitem{7} Dutoit, T.: High-quality Text-to-Speech Synthesis: An Overview. J. of Electrical and Electronics Engineering, Special Issue on Speech Recognition and Synthesis, vol. 17(1), pp. 25-37 (1997)
\bibitem{8}	Gros, J., Pavešić, N., Mihelič, F.: Text-to-Speech Synthesis: A Complete System for the Slovenian language. Journal CIT, Vol. 5(1), pp. 11-19 (1997)
\bibitem{9}	Taylor, P., Black, A., Caley, R.: The Architecture of the Festival Speech Synthesis System. ESCA Workshop in Speech Synthesis, Australia, pp. 147-151 (1998)
\bibitem{10}	Reichel, U. D., Pfitzinger, H. R.: Text Preprocessing for Speech Synthesis. Univ. of Munich (2006)
\bibitem{11} Schwartz, R. L., et al.: Learning Perl. O'Reilly (2005) 
\bibitem{12}	Sproat, R., et al.: Normalization of Non-standard Words. Computer Speech \& Language. Vol.5. pp. 287-333 (2001)
\bibitem{13}	Sproat, R.: Lightly Supervised Learning of Text Normalization: Russian Number Names, IEEE Workshop on Spoken Language Technology, Berkeley, CA (2010)
\bibitem{14}	Načinović, L. et al.: Grapheme-to-Phoneme Conversion for Croatian Speech Synthesis. MIPRO 2009, CIS,  pp. 18-323 (2009)
\bibitem{15}	Pobar, M., Martinčić-Ipšić, S., Ipšić, I.: Text-to-Speech Synthesis: A Prototype System for the Croatian Language. Engineering Review, Vol. 28(2), pp.31-44 (2008) 
\bibitem{16} Kanis, J., Zelinka, J., Müller, L.: Automatic Numbers Normalization in Inflectional Languages. SPECOM 2005, Moscow , pp. 663-666 (2005)
\end{thebibliography}
\end{document}